\pdfoutput=1

\documentclass[11pt]{article}

\usepackage[]{ACL2023}
\usepackage{amsmath}
\usepackage{times}
\usepackage{latexsym}
\usepackage{newtxtext} 
\usepackage{newtxmath} 
\usepackage[T1]{fontenc}

\usepackage[utf8]{inputenc}
\usepackage{booktabs}
\usepackage{anyfontsize}
\usepackage{microtype}
\usepackage{tcolorbox}
\usepackage{inconsolata}
\usepackage{soul}
\usepackage{multirow}
\usepackage{epstopdf}
\usepackage{hyperref}
\usepackage{listings}
\usepackage{fancybox}
\usepackage{graphicx}
\usepackage{subcaption}
\usepackage{paralist}
\usepackage{ragged2e}
\usepackage{enumitem}
\usepackage{tcolorbox}
\makeatletter
\usepackage{xcolor}
\usepackage{algpseudocode}

\usepackage{color}
\usepackage{xcolor,colortbl}
\usepackage{balance}
\usepackage{threeparttable}

\usepackage[framemethod=TikZ]{mdframed}
\usepackage{tcolorbox}
\makeatletter
\newcommand{\mybox}[1]{%
  \setbox0=\hbox{#1}%
  \setlength{\@tempdima}{\dimexpr\wd0+13pt}%
  \begin{tcolorbox}[boxrule=0.5pt, colback=white, arc=4pt,
      left=6pt,right=6pt,top=6pt,bottom=6pt,boxsep=0pt]
    #1
  \end{tcolorbox}
}
\usepackage{tikz}
\usepackage{pgfplots}
\usetikzlibrary{pgfplots.statistics,calc}
\usepackage{color}
\usepackage{array}
\usepackage{amsmath}
\usepackage{centernot}
\usepackage{xspace}
\usepackage{url}
\usepackage{verbatim}
\usepackage{wrapfig}
\usepackage{tabularx}
\usepackage{bbding}
\usepackage{pifont}
\usepackage{wasysym}
\usepackage{cleveref}
\usepackage{makecell}

\makeatletter  
\newif\if@restonecol  
\makeatother  
\usepackage[linesnumbered,ruled,vlined]{algorithm2e}
\usepackage{algpseudocode}  
\newcommand{\tool}{\textit{Patcher}}
\title{ Repairing Catastrophic-Neglect in Text-to-Image Diffusion Models via Attention-Guided Feature Enhancement}

\author{
\fontsize{10pt}{6pt}\selectfont
  Zhiyuan Chang\textsuperscript{\normalfont 
 1,2,3}   \hspace{0.5cm}
  Mingyang Li\textsuperscript{\normalfont 
 1,2,3}\Thanks{ Corresponding authors}  \hspace{0.5cm}
  Junjie Wang\textsuperscript{\normalfont  1,2,3} \\
  \fontsize{10pt}{6pt}\selectfont
  \textbf{Yi Liu}\textsuperscript{4} \hspace{0.5cm}
  \textbf{Qing Wang}\textsuperscript{1,2,3}\footnotemark[1]  \hspace{0.5cm}
  \textbf{Yang Liu}\textsuperscript{4} \\
  \fontsize{10pt}{6pt}\selectfont
  \textsuperscript{1}State Key Laboratory of Intelligent Game, Beijing, China \\
\fontsize{10pt}{6pt}\selectfont
 \textsuperscript{2}Science and Technology on Integrated Information System Laboratory, \\
  \fontsize{10pt}{6pt}\selectfont
Institute of Software Chinese Academy of Sciences, Beijing, China \\
  \fontsize{10pt}{6pt}\selectfont
  \textsuperscript{3}University of Chinese Academy of Sciences
  \hspace{0.3cm}
  \textsuperscript{4}Nanyang Technological University \\
  \fontsize{10pt}{6pt}\selectfont
  \textit{\{zhiyuan2019, mingyang2017, junjie, wq\}@iscas.ac.cn, 
  yi009@e.ntu.edu.sg, yangliu@ntu.edu.sg}
}

\begin{document}
 \maketitle
\begin{abstract}
Text-to-Image Diffusion Models (T2I DMs) have garnered significant attention for their ability to generate high-quality images from textual descriptions.
However, these models often produce images that do not fully align with the input prompts, resulting in semantic inconsistencies.
The most prominent issue among these semantic inconsistencies is catastrophic-neglect, where the images generated by T2I DMs miss key objects mentioned in the prompt.
We first conduct an empirical study on this issue, exploring the prevalence of catastrophic-neglect, potential mitigation strategies with feature enhancement, and the insights gained.
Guided by the empirical findings, we propose an automated repair approach named {\tool} to address catastrophic-neglect in T2I DMs.
Specifically, {\tool} first determines whether there are any neglected objects in the prompt, and then applies attention-guided feature enhancement to these neglected objects, resulting in a repaired prompt.
Experimental results on three versions of Stable Diffusion demonstrate that {\tool} effectively repairs the issue of catastrophic-neglect, achieving 10.1\%-16.3\% higher Correct Rate in image generation compared to baselines. 

\end{abstract}

\section{Introduction}
\label{sec:introduction}
Text-to-Image Diffusion Models (T2I DMs) \cite{RombachBLEO22,SahariaCSLWDGLA22,dell2} have gained widespread attention in recent years due to their remarkable ability to generate images from textual descriptions (i.e. prompt). 
However, it has been demonstrated that the image generated by T2I DMs may not strictly adhere to the description of the input prompt, leading to inconsistencies in the semantics.

To this end, many approaches have been proposed to enhance the generation quality through  inference process optimization~\cite{LiuLDTT22,FengHFJANBWW23,CheferAVWC23} and hand-crafted prompt writing guidelines~\cite{LiuC22a,oppenlaender2022taxonomy}.
The former requires modifications to the model structure or parameters, which is difficult for users to perform.
Although the latter is relatively easier to implement, it requires a significant amount of manual effort and suffers poor scalability.
Recently, \citet{HaoC0W23} also proposed a method to enhance the quality of generated images by automating the refinement of user-inputted prompts.



\begin{figure}[t!]
\centering
\setlength{\abovecaptionskip}{5pt}   
  \setlength{\belowcaptionskip}{0pt} 
\includegraphics[width=7cm,height=7.1cm]{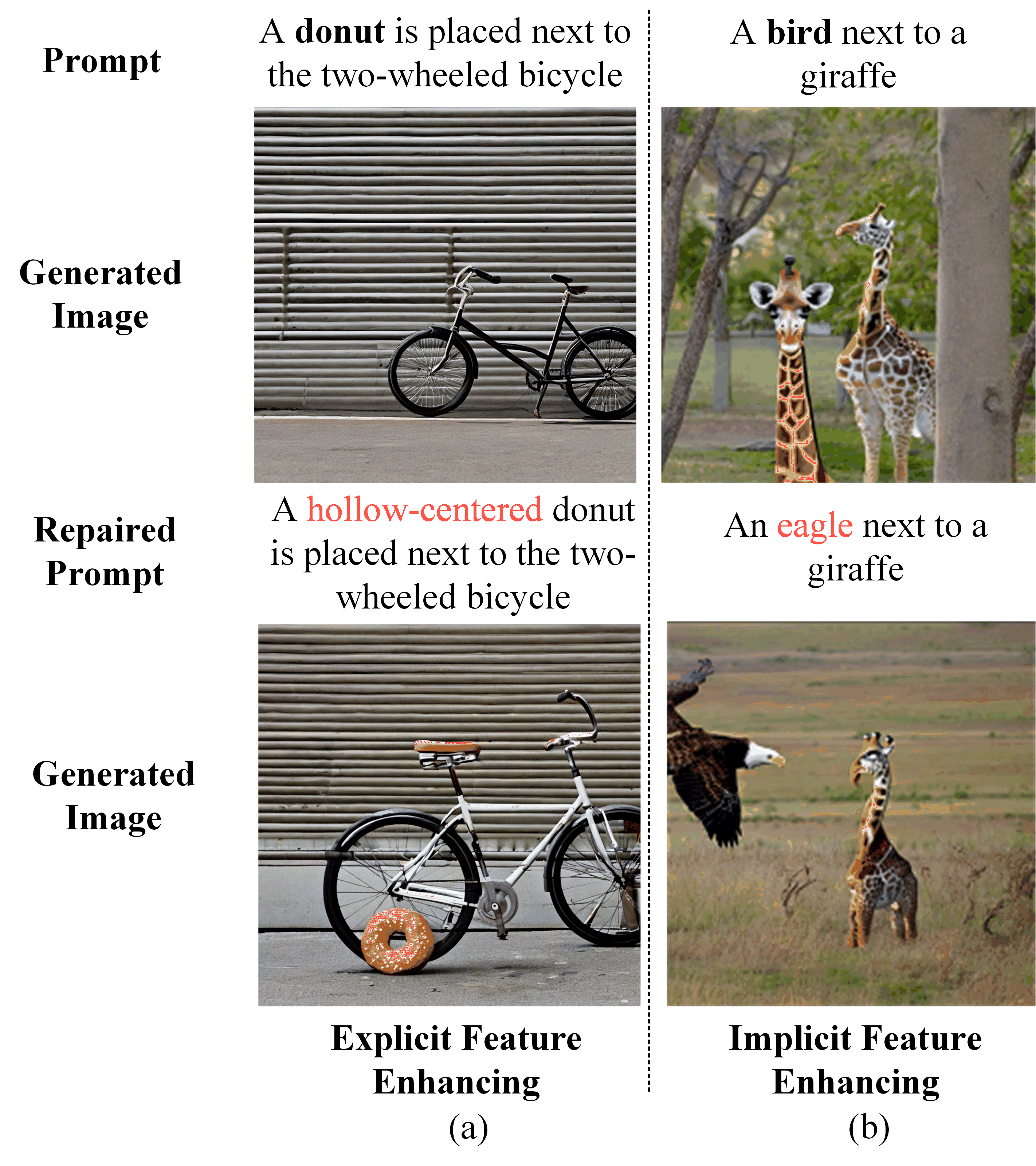}
\caption{
Examples of catastrophic neglect in the generated images by T2I DMs, and the enhancement of explicit and implicit features.
}
\label{fig:motivation}
\end{figure}

According to previous study~\cite{CheferAVWC23}, one of the most prominent issues in semantic consistency is the \textbf{catastrophic-neglect}, i.e., the images generated by T2I DMs often miss some of the key objects mentioned in the textual prompts.
This issue is particularly prevalent when a prompt involves multiple objects.
Figure \ref{fig:motivation} demonstrates two illustrative cases where one of the two objects is neglected by T2I DMs.
In Figure \ref{fig:motivation} (a), we notice that the object ``bicycle'' in prompt is described with the explicit feature ``two-wheeled'' while ``donut'' is not.
We try to craft prompts to repair the issue, and results reveal that by adding a specific explicit feature to the ``donut'' (e.g., ``hollow-centered''), the catastrophic-neglect issue can be resolved. 
Furthermore, as the feature is added, the attention difference between the two mentioned objects (i.e., ``bicycle'' and ``donut'') is reduced according to the explainable tool \cite{TangLPJYKSLT23}.
It seems that reduction in attention difference can potentially indicate the T2I DMs put more balanced attention towards the two involved objects, resulting both of them can be successfully generated. 
In Figure \ref{fig:motivation} (b), we notice that the object ``bird'' in the prompt is a more general concept with fewer implicit features compared with the concept ``giraffe'', according to the hierarchical structure in WordNet ~\cite{Miller95}. 
Taken in this sense, we can successfully repair the issue through using more imageable concept (such as ``eagle'' ) to replace ``bird'' in the prompt, and the attention difference between two mentioned objects (i.e., ``eagle'' and ``giraffe'') is also reduced. 
Motivational study in Section \ref{sec:preliminary}  provides more details.

Motivated by the above analysis, we assume the attention difference can guide the mitigation of catastrophic-neglect issue, and this can be achieved  through enhancing objects with specific features (i.e., explicit features) or using more imageable concepts (i.e., implicit features) to balance the attention among involved objects in the prompt. 

Therefore, this paper proposes an automatic repair approach named {\tool} to address catastrophic-neglect in T2I DMs, guided by the attention difference among objects of input prompt. 
Specifically, {\tool} first parses the original prompt and identifies the objects neglected by the T2I DMs. 
Then, guided by the difference of attention scores, {\tool} produces the repaired prompt via enhancing explicit feature (achieved by asking LLMs for suitable modifiers) and implicit features (realized by hyponym substitution using WordNet), and re-determined whether there are still neglected objects in the generated image. 

Experimental results demonstrate that {\tool} effectively repairs the issue of catastrophic-neglect in T2I DMs, achieving 10.1\%-16.3\% higher Correct Rate in image generation compared to baselines, as tested on Stable-Diffusion V1.4, V1.5, and V2.1 models.
Additionally, ablation study shows that both explicit and implicit feature enhancing in {\tool} contribute to resolving the catastrophic-neglect issue in T2I DMs. We provide the public reproduction package\footnote{https://github.com/lsplx/patcher}.

\section{Motivation}
\label{sec:preliminary}

To better understand catastrophic-neglect and guide the design of the automated repair approach, we conduct the empirical analysis from three aspects, i.e., their prevalence across prompts with different number of objects, potential mitigation strategies based on feature enhancement, and corresponding insights into the effectiveness of feature enhancement.

\subsection{Issue Prevalence}
\label{sec:error_analysis}
On the one side, we investigate the error rate of T2I DMs in handling prompts involving different numbers of objects through manual evaluation. 
On the other side, we explore the proportion of catastrophic-neglect among all errors.
First, we construct three datasets containing single-object, double-object and triple-object prompts respectively.
For the single-object prompts, we reuse the 80 object descriptions from different semantic categories in MSCOCO dataset~\cite{LinMBHPRDZ14}
Based on these single-object prompts, we synthesize new prompts containing two or three objects using GPT-3.5 by adding essential conjunctions, adverbs or interactions
, aiming to generate inputs for T2I DMs that conform to human expressions\footnote{The size of the dataset is described in Sections \ref{sec:dataset}.}.

We then input the single-object prompts and multi-object prompts into Stable Diffusion V2.1, a state-of-the-art T2I DM, and manually evaluate the proportion of incorrectly generated images that are not consistent with the prompt (i.e. Error Rate).

    

The evaluation results show that the Error Rate significantly increases (2.5\%->50.4\%->86.0\%) with the numbers of objects in the prompt.
Furthermore, for the prompts with single, double and triple object, catastrophic-neglect issue accounts for 100\%, 93.4\%, and 94.0\% of all the incorrectly generated images. 
The remaining incorrectly images are those where the features of multiple objects are blended into a single object.
In general, when faced with multi-object prompts, the T2I DM is prone to generating incorrect images, with catastrophic-neglect being the most severe issue in such scenario.

\subsection{Issue Mitigation via Feature Enhancement}
\label{sec:feature_enhance}

Section \ref{sec:introduction} has illustratively demonstrated that the imbalance of explicit/implicit features carried by objects in the prompts may lead to the catastrophic-neglect. 
This section tries to craft the prompts with the idea of adding explicit or implicit features to those neglected objects to investigate whether the issue could be mitigated statistically. 
Specifically, we apply feature enhancement to double-object and triple-object datasets (Constructed in Section \ref{sec:error_analysis} with 4041 prompts).
First, we manually add explicit features to the neglected objects.
These features enhance the physical appearance of the original objects without altering the semantic meaning of the original prompts. 
As shown in Figure \ref{fig:motivation}, the neglected object ``donut'' was enhanced with the feature ``hollow-centered''.

Second, we enhance the prompts using implicit features.
We manually replace the description of the neglected object with its hyponym with help of WordNet, which denotes a specific concept compared to the original object~\cite{Miller95}.
As shown in Figure \ref{fig:motivation}, we replaced ``bird'' with ``eagle'' to obtain the repaired prompt. 
The evaluation results show that, compared to the Error Rate before feature enhancement, manually constructed explicit and implicit features reduce Stable Diffusion’s Error Rate by 26.9\% and 24.6\%, respectively.




\subsection{Explanation for Feature Enhancement}
\label{sec:Interpretability}
To explore the reasons behind feature enhancement, we use the attention explainability tool (DAAM)~\cite{TangLPJYKSLT23} to investigate whether the attention differences between multiple objects change before and after feature enhancement.
Given a specific token from the input prompt, DAAM aggregates the T2I DM’s cross-attention values across layers to obtain its \textbf{attention score}.
The attention score of each token represents the token’s importance in the image generation process.
The \textbf{attention difference} indicates the disparity in the T2I DM's attention score to different object tokens.
We assume that reducing the attention difference between multiple objects can help the T2I DM more evenly focus on the features of each object and generate them correctly.
For double-object prompts, we compute the absolute difference in attention scores between the two objects.
For prompts with triple object, we first calculate the pairwise differences in attention scores and then average them.
We use the prompts that generates incorrectly images from multi-objects prompts constructed in Section \ref{sec:error_analysis} and the repaired prompts manually constructed in Section \ref{sec:feature_enhance}.

\begin{table}[t]
 \setlength{\abovecaptionskip}{5pt}   
  \setlength{\belowcaptionskip}{0pt}
  \caption{The attention difference between multiple objects before and after using explicit and implicit features. 
  `Correct' and `Wrong' respectively indicates the results of the newly generated images after adding the features.
  }
  \label{tab:attetion_difference}
  
\resizebox{0.48\textwidth }{!}{
\begin{threeparttable}
\begin{tabular}{c|ccccc}
\toprule

  \multirow{2}{*}{\textbf{Strategy}} & 

  \multicolumn{3}{c}{\textbf{Correct}}&
  \multicolumn{2}{c}{\textbf{Wrong}}
  \\     \cline{2-3} \cline{5-6}

     &

  \multicolumn{1}{c}{\textit{\textbf{Before}}} &

  \textit{\textbf{After}} &
  &

  \multicolumn{1}{c}{\textit{\textbf{Before}}} &

  \textit{\textbf{After}} \\

 \midrule

  \textit{Explicit Feature} &

658&

  \multicolumn{1}{c}{232}& &
  808&

  \multicolumn{1}{c}{887}

  \\

   \midrule

  \textit{Implicit Feature} &


   934&

  \multicolumn{1}{c}{437}& &
  713&

  \multicolumn{1}{c}{1442} \\

  \bottomrule
\end{tabular}
  \end{threeparttable}
}
\end{table}

The result is shown in Table \ref{tab:attetion_difference}. The attention difference between multiple objects significantly decreases for prompts that correctly generate images after enhancing explicit or implicit features. Besides, this reduction in attention difference accounts for 80.9\% of the correctly generated images.
In contrast, for prompts that still generate incorrect images, the attention difference increases. Moreover, the reduction in attention difference accounts for 29.0\% of these incorrect generated images.
This indicates that features reducing the attention difference between objects are more effective in repairing the catastrophic-neglect in the T2I DM.
\section{Methodology}
\label{sec:approach}

\begin{figure*}[htbp]
  \setlength{\abovecaptionskip}{5pt}   
  \setlength{\belowcaptionskip}{0pt} \center{\includegraphics[width=\linewidth]{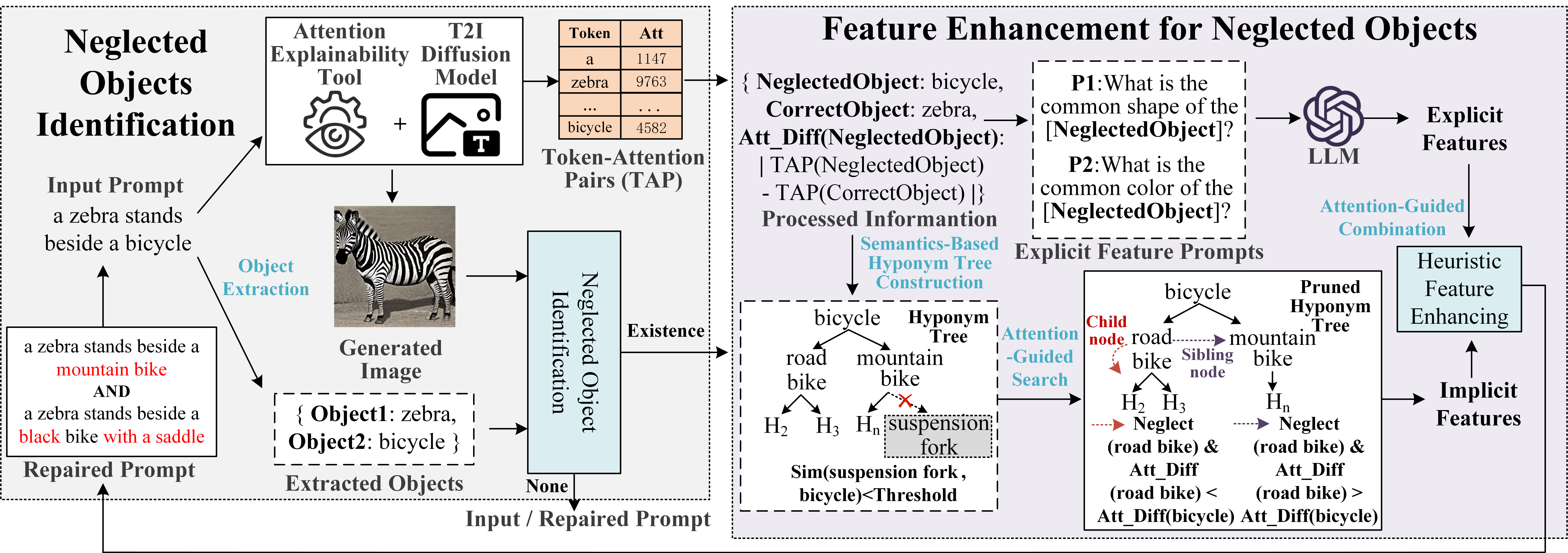}}
    \caption{
     The overview of {\tool}. The procedure in the dashed box is executed only the first time.
     } 
    \label{fig:artifacture}    
\end{figure*}

Figure \ref{fig:artifacture} shows the overview of {\tool}.
{\tool} consists of two stages: (1) \textbf{Neglected Objects Identification} would determine whether the T2I DM neglect any objects in the input prompt; (2) \textbf{Feature Enhancement for Neglected Objects} would enhance explicit and implicit features for neglected objects and construct the repaired prompt.

\subsection{Neglected Objects Identification}
\label{subsec_method_neglect}
To identify the neglected objects, {\tool} first extracts the objects from the input prompt.
Specifically, {\tool} first parses the textual descriptions into a dependency tree using a transformer-based~\cite{VaswaniSPUJGKP17} language model\footnote{https://huggingface.co/spacy/en\_core\_web\_trf}.
It then extracts noun phrases from this tree as the object entities.
In the meanwhile, {\tool} employs DAAM to obtain the attention scores and produces the token-attention pairs (TAP) for each token in the prompt description, which will be utilized in Section \ref{sec:method_feature_enhance} to guide the feature enhancement.

After that, {\tool} calculates the similarities of each extracted object entities and generated images by Clipscore~\cite{RadfordKHRGASAM21}.
Due to the presence of corresponding visual features when the object is in the image and their absence when it is not, there is a significant difference in similarity between the two scenarios, 
{\tool} sets a threshold based on the empirical study to determine whether an object is neglected in the image. 
If the similarity between the object and the image is below the threshold, we consider the object to be neglected by the T2I model.
Conversely, we consider the object to be correctly generated by the T2I model.
If there are no neglected objects in the prompt, output the current prompts; otherwise,
{\tool} sends the prompt into the following stage for repair.

\subsection{Feature Enhancement for Neglected Objects}
\label{sec:method_feature_enhance}
After the first stage, {\tool} derives a set of neglected objects and a set of correctly identified objects. Recall that it also obtains the attention scores for each token in the first stage.
Typically, an object contains a single token; if it contains multiple tokens, {\tool} calculates the average of the attention scores for these tokens.
In this way, we obtain the attention score for each neglected object and correct object.
We then calculate the differences in attention scores between neglected objects and correct objects.

Specifically, it first calculates the pairwise differences between attention scores of objects from the neglected and correct object sets, and then averages these differences.
This provides a comprehensive measure of how uniformly the T2I DM's attention is distributed between two set of objects.
The calculation process is shown in Equation~\ref{eq:average_difference}, where \(O_i\) denotes the attention score corresponding to the i-th object from the neglected object set, and \(O_j\) denotes the attention score corresponding to the j-th object from the correct object set. 

\begin{equation}
\label{eq:average_difference}
\small
\text{Att\_Diff} = \frac{1}{|N||C|} \sum_{O_i \in N} \sum_{O_j \in C} |O_i - O_j|
\end{equation}

Next, {\tool} employs two repair strategies: 1) \textit{Explicit Feature Enhancing}, which is used to obtain the physical features of the neglected objects; 2) \textit{Implicit Feature Enhancing}, which is used to obtain hyponyms of the neglected objects guided by the attention difference.
With the two strategies, {\tool} simultaneously generates explicit and implicit features, each forming a repaired prompt, which together constitute two
repaired prompts to determine whether there are neglected objects in them. 
Following introduces the prompt repair process with the two strategies respectively.


\subsubsection{Explicit Feature Enhancement}

From the explicit perspective, objects' features are enhanced from two aspects, i.e., shape and color, leveraging the LLM's powerful understanding of the general knowledge~\cite{chang2024survey} with the carefully designed prompt (See Appendix \ref{sec:appendix_prompt} for specific details).

The prompt consists of three parts: 1) the specific question, which directly asks the LLM about the core objective, 
2) the output guidelines, which constrain the format of the model's output and guide it to produce diverse responses, and 3) the example, which helps the LLM understand the question and produce the response expected by the users.
As shown in Figure \ref{fig:artifacture}, {\tool} inputs the explicit feature prompts into the LLM\footnote{The LLM is GPT-3.5}, which in return provides a variety set of explicit features.
For each explicit feature, {\tool} replaces the description of the neglected object in the original prompt with an enhanced description containing the object and its explicit feature, generating a candidate prompt.
{\tool} iteratively queries the T2I models with the candidate prompts until no neglected objects (determined with the strategy in Section \ref{subsec_method_neglect}) or reaching the maximum iteration number (set as 4 in our study).
If all color and shape explicit features fail to make the neglected object visible in the image, {\tool} selects the feature with the smallest attention difference from both candidate sets of color and shape, then combines them to generate the final repaired prompt.


\subsubsection{Implicit Feature Enhancement}
To obtain the hyponyms of a neglected object, {\tool} uses Natural Language Processing tool \cite{nltk} to search all hyponyms of the object, i.e., including the direct hyponyms and those indirect hyponyms, recursively, until no further hyponyms are found.
As shown in the hyponym tree in Figure \ref{fig:artifacture}, {\tool}  constructs a hyponym tree for ``bicycle'', where the child node ``mountain bike'' is a direct hyponyms of ``bicycle''.
For nodes at the same hierarchical level, such as ``mountain bike'' and ``road bike'',
their conceptual levels are similar, making them sibling nodes.
Besides, the child nodes of ``mountain bike'' are indirect hyponyms of ``bicycle''.
Among these, some indirect hyponyms such as ``Suspension Fork'' have already deviated from the original semantic concept of the root node ``bicycle'', which could not help the T2I DM generate correct original object.
To mitigate this issue, {\tool} performs semantic-based pruning for the hyponym tree.
Specifically, by traversing each child node of the hyponym tree using breadth-first search, {\tool} maps the textual representation of the current node object  and neglected object into a vector space using a language model \cite{brown2020language}, then computes the cosine similarity between them. 
If the similarity is below a certain threshold, {\tool} prunes the current node and its children.

After that, {\tool} performs an attention-guided search on the pruned hyponym tree, as detailed in Algorithm \ref{alg:atten_search}.
For each node, {\tool} first replaces the neglected object in the original prompt with the hyponym represented by that node (Line 2-4).
Then, input the generated repaired prompt into the Neglected Objects Identification Stage to judge whether the neglected object still exists.
If there are no neglected objects, output the repaired prompt (Line 5-8); otherwise, proceed with the attention-guided search (Line 9-15).
Specifically, {\tool} calculates the attention difference between the replaced hyponym and the correct objects in the repaired prompt, then compares it with the original attention difference between neglected object and correct objects.
Considering that child nodes contains more implicit features compared to sibling nodes, if the attention difference is reduced, {\tool} continues the search with the child nodes of the current node; otherwise, search its sibling nodes.

\begin{algorithm}[t]
\caption{Attention-Guided Search}
\footnotesize
\label{alg:atten_search}
\KwIn{Hyponym tree $T$ with root node (Neglected Object) $NO$, original prompt $P$, and Att\_Diff(NO)}
Initialize a queue $Q$ with tree $T$ and enqueue the root node $NO$\;
\While{$Q$ is not empty}{
    $node = Q.\text{dequeue}()$\;
    $\text{repaired prompt} = \text{replace}(P, NO, node)$\;
    $judgment = \text{Judge}(\text{repaired prompt})$\;
    \If{$judgment$ is None}{
        \textbf{output} $\text{repaired prompt}$\;
        \Return\;
    }
    \Else{
        \If{$Att\_Diff(node) < Att\_Diff(NO)$}{
            \For{each child $c$ of $node$}{
                $Q.\text{enqueue}(c)$\;
            }
        }
        \Else{
        \For{each sibling $s$ of $node$}{
        $Q.\text{enqueue}(s)$\;}
    }
    }
    
}
\textbf{output} ``No correct node found''\;
\end{algorithm}

During the process of explicit and implicit feature enhancement, if a correct image is generated, the corresponding repaired prompt is returned.
Otherwise, the prompt that achieves the minimum attention difference is returned.

\section{Experimental Setup}
\label{sec:experiment}


\subsection{Datasets}
\label{sec:dataset}

For experimental evaluation, we first introduce the popularly used datasets constructed by HILA et al. \cite{CheferAVWC23} for T2I task.
Given that publicly available datasets only involve prompts with double objects combined by an ``and'' relationship, we further based on some of the 80 single objects in MSCOCO with the help of LLM (same as the datasets in Section \ref{sec:error_analysis}).
Followings introduces the details of the datasets.


\begin{itemize}
    \item Template-Based Pairs (TBP): It is the public dataset constructed by HILA et al.~\cite{CheferAVWC23} used for T2I task.
    All the prompts in the dataset contain two objects that are constructed by three templates, i.e.,
    ``a [animalA] and a [animalB]'', ``a [animal] and a [color][object]'', and ``a [colorA][objectA] and a [colorB][objectB]''. 
    The placeholders in the templates are filled with 12 types of animals, 12 objects and 11 colors.
    \item Two/Three-Object Prompts (TwOP/ThreeOP): The detailed construction of our two datasets can be found in Section \ref{sec:error_analysis}. 
    The constructed datasets contain 3,160 prompts with two objects and the same number of prompts with three objects.
\end{itemize}

\subsection{Subject Models}
To investigate the performance of {\tool} in repairing catastrophic-neglect issue.
we introduce three T2I DMs (Stable Diffusion V1.4 (SD V1.4), Stable Diffusion V1.5 (SD V1.5), and Stable Diffusion V2.1 (SD V2.1)) for their wide adoption in community. All models are run on a 3090 GPU with 24GB of VRAM.

\subsection{Evaluation Metric and Measurement Method}
\label{sec:metric}

We adopt two evaluation metrics.
\begin{itemize}
    
    \item \textit{CLIPScore}: it measures the similarity between the input prompt and generated image, and is used in many previous studies~\cite{HaoC0W23,CheferAVWC23}.
    However, it serves as a weaker indication of image-text similarity in T2I task, as correctness of generated images cannot be absolutely determined directly based on the magnitude of the value.

    \item \textit{Correct Rate (CR)}: the percentage of correctly generated images out of all generated images.
    Compared to CLIPScore, CR is a direct measurement indicating whether a generated image is correct.
    For an image generated by T2I models, we manually judge whether it is correct by a annotation team consisting of one senior researcher and two Ph.D students.
    If more than half of the members perceive the generated image to be semantically consistent with the input prompt, we consider it as a correctly generated one.
    
\end{itemize}

\subsection{Baselines}
Our baselines include approaches based on prompt optimization (Promptist) and inference process optimization (AE).
Besides above two baselines specific for T2I DMs, we have also specifically established a baseline that iteratively refines the output results through iterative queries (LR), which is a commonly-used strategy for performance improvement in the LLM context~\cite{Patrick2023,Anay2023}.

    \textit{Promptist~\cite{HaoC0W23}}
    is the state-of-the-art approach to improve the generation quality of T2I DMs via prompt optimization.
    It first performs supervised fine-tuning with a pretrained language model on a small collection of manually engineered prompts. Then it defines a reward function that encourages the T2I DM to generate more aesthetically pleasing images while preserving the original prompt intentions. After that, it uses reinforcement learning with the reward function to further boosts performance of the fine-tuned model.
    
    \textit{Attend-and-Excite (AE)~\cite{CheferAVWC23}} 
    is the state-of-the-art approach specific for catastrophic-neglect in T2I DMs via inference process optimization.
    Specifically, it adds an attention guidance mechanism during the model's inference stage to enhance the cross-attention units.
    This mechanism ensures that the model attends to all object tokens in the text prompts and boosts their activations, thereby encouraging the model to generate all objects described in the text prompts.
    However, AE requires prior knowledge of the positions of object tokens in the original prompts. 
    For the input prompts, we use the object extraction method in {\tool} to identify and return the positions of the objects within the prompts. 
    Finally, the prompts, along with the positional information of the objects, are fed into the T2I DM enhanced by AE to generate optimized images.

    \textit{LLM-Repair (LR)}
    improves the quality of the generated images by the iterative query strategy that is commonly employed in practice to improve the outputs in the LLM context. 
    Specifically, with the original prompt, LR first identifies the neglected objects in the generated employing the first stage in {\tool}.
    After that, LR leverages GPT-3.5 to produce the new prompt describing the details of the neglected objects and asking for the T2I model to mitigate the catastrophic-neglect as much as possible in the next iteration (the prompt templates in shown in Appendix \ref{sec:LLMRepair}).
    Then, LR iteratively query the T2I models until no object is identified as neglected one or reaching the maximum iteration number (set as 8 in our study). 
    

\section{Results}
\label{sec:result}

We designed two sets of experiments to explore the performance of {\tool} in repairing catastrophic-neglect: the effectiveness of {\tool} and the ablation study within {\tool}.

\subsection{Effectiveness of {\tool}}



Table \ref{tab:RQ1_performance} shows the effectiveness of {\tool} and baselines in CR and CLIPScore.
The column ``Original'' represents the quality of the images generated by different T2I DMs with the original prompts in the three datasets.
The last four columns show the quality of the generated images after repair for three baselines and {\tool} respectively.
From the perspective of CR, {\tool} achieves the best performance across all T2I models under testing and datasets, surpassing the baselines of 10.1\%-16.3\%.
Especially on the last two datasets, TwOP and ThreeOP with more complex inter-object relationships or a greater number of objects, {\tool} shows a more substantial improvement (31.8\% higher than the original prompts and 12.4\%-21.9\% higher than the three baselines).

Compared to Promptist, {\tool} achieves an CR improvement of 16.3\%. 
Promptist automates the addition of modifiers at the end of the input prompts, such as ``highly detailed'', ``masterpiece'', or ``sharp focus'', to enhance the quality of the generated images.
Adding such modifiers could help the T2I DM focus more on depicting the overall semantics of the prompt.
However, in cases where there are significant feature differences between multiple objects, enhancing the T2I DM's focus on the entire sentence of the prompt could not effectively narrow the attention difference between different objects.
It still requires the addition of appropriate modifiers to objects with weaker features.
As for AE, it optimizes the inference process within T2I DMs rather than the prompts, which is supposed to be effective in principle but more difficult for end users to perform.
However, {\tool} still achieves superior performance compared to AE, with a CR improvement of 10.1\%.
As for LR, similar to {\tool}, multiple attempts are needed to repair the prompts.
Statistical analysis shows that LR requires an average of 5.7 attempts to correctly repair an image, whereas {\tool} requires only 2.3 attempts.
Additionally, {\tool}'s CR exceeds LR by 14.2\%, demonstrating the effectiveness of feature enhancement.
The result also implies that if lacking guidance on feature enhancement, relying solely on the intrinsic capabilities of T2I MDs makes it difficult to effectively improve the accuracy of generated images.

\begin{table}[t]
\setlength{\abovecaptionskip}{5pt}   
  \setlength{\belowcaptionskip}{0pt}
\Huge
  \caption{The Correct Rate (CR) and the ClIPScore of the original prompts, {\tool} and baselines.}
  \label{tab:RQ1_performance}
\resizebox{0.5\textwidth }{!}{
\begin{threeparttable}
\begin{tabular}{ccc|ccccc}
\toprule
\multirow{1}{*}{\textbf{Dataset}} &
\multirow{1}{*}{\textbf{Model}} &
  \multirow{1}{*}{\textbf{Metric}} & 
  \multirow{1}{*}{\textbf{Original}} &
  \multirow{1}{*}{\textbf{LR}} &
  \multirow{1}{*}{\textbf{Promptist}} &
   \multirow{1}{*}{\textbf{AE}} &
   \multirow{1}{*}{\textbf{{\tool}}} \\     

\midrule
\multirow{6}{*}{TBP} &
\multirow{2}{*}{{SD V1.4}} &
  \textit{CR} &

  \multicolumn{1}{c}{61.4\%} & 
  \multicolumn{1}{c}{75.0\%} & 
  \multicolumn{1}{c}{78.9\%} & 
  \multicolumn{1}{c}{83.6\%} & 
  \multicolumn{1}{c}{\textbf{89.8\%}}  \\ 

& &\textit{CLIPScore} &

  \multicolumn{1}{c}{32.0\%} & 
  \multicolumn{1}{c}{32.2\%} & 
  \multicolumn{1}{c}{32.2\%} & 
  \multicolumn{1}{c}{32.6\%} & 
  \multicolumn{1}{c}{\textbf{32.7\%}} \\

\cline{2-8}
  \rule{0pt}{2.6ex}

& \multirow{2}{*}{{SD V1.5}} &
  \textit{CR} &
\multicolumn{1}{c}{55.1\%} & 
\multicolumn{1}{c}{76.1\%} & 
  \multicolumn{1}{c}{78.2\%} & 
  \multicolumn{1}{c}{79.3\%} & 
  \multicolumn{1}{c}{\textbf{88.0\%}} \\ 

  &  & \textit{CLIPScore} &
\multicolumn{1}{c}{31.8\%} & 
\multicolumn{1}{c}{32.0\%} & 
  \multicolumn{1}{c}{32.1\%} & 
  \multicolumn{1}{c}{\textbf{32.3\%}} & 
  \multicolumn{1}{c}{\textbf{32.3\%}}  \\ 
\cline{2-8}
  \rule{0pt}{2.6ex}

& \multirow{2}{*}{{SD V2.1}} &
  \textit{CR} &
\multicolumn{1}{c}{72.4\%} & 
\multicolumn{1}{c}{84.4\%} & 
  \multicolumn{1}{c}{81.1\%} & 
  \multicolumn{1}{c}{85.4\%} & 
  \multicolumn{1}{c}{\textbf{96.0\%}} \\ 

  &  & \textit{CLIPScore} &
\multicolumn{1}{c}{32.8\%} & 
\multicolumn{1}{c}{33.0\%} & 
  \multicolumn{1}{c}{32.7\%} & 
  \multicolumn{1}{c}{33.2\%} & 
  \multicolumn{1}{c}{\textbf{33.4\%}}  \\

  \midrule

 \multirow{6}{*}{TwOP} &
 \multirow{2}{*}{{SD V1.4}} &
  \textit{CR} &
\multicolumn{1}{c}{45.6\%} & 
\multicolumn{1}{c}{63.6\%} & 
  \multicolumn{1}{c}{53.8\%} & 
  \multicolumn{1}{c}{63.2\%} & 
  \multicolumn{1}{c}{\textbf{77.8\%}} \\ 

  &  & \textit{CLIPScore} &
\multicolumn{1}{c}{30.3\%} & 
\multicolumn{1}{c}{30.5\%} & 
  \multicolumn{1}{c}{29.5\%} & 
  \multicolumn{1}{c}{30.6\%} & 
  \multicolumn{1}{c}{\textbf{30.7\%}}  \\ 

\cline{2-8}
  \rule{0pt}{2.6ex}
  
& \multirow{2}{*}{{SD V1.5}} &
  \textit{CR} &
\multicolumn{1}{c}{45.8\%} & 
\multicolumn{1}{c}{67.9\%} & 
  \multicolumn{1}{c}{56.2\%} & 
  \multicolumn{1}{c}{68.2\%} & 
  \multicolumn{1}{c}{\textbf{78.0\%}} \\ 

  &  & \textit{CLIPScore} &
\multicolumn{1}{c}{30.1\%} & 
\multicolumn{1}{c}{30.6\%} & 
  \multicolumn{1}{c}{29.4\%} & 
  \multicolumn{1}{c}{\textbf{30.7\%}} & 
  \multicolumn{1}{c}{\textbf{30.7\%}}  \\ 

\cline{2-8}
  \rule{0pt}{2.6ex}

& \multirow{2}{*}{{SD V2.1}} &
  \textit{CR} &
\multicolumn{1}{c}{49.6\%} & 
\multicolumn{1}{c}{69.1\%} & 
  \multicolumn{1}{c}{63.8\%} & 
  \multicolumn{1}{c}{69.4\%} & 
  \multicolumn{1}{c}{\textbf{80.2\%}} \\ 

  &  & \textit{CLIPScore} &
\multicolumn{1}{c}{30.5\%} & 
\multicolumn{1}{c}{30.7\%} & 
  \multicolumn{1}{c}{30.1\%} & 
  \multicolumn{1}{c}{\textbf{30.9\%}} & 
  \multicolumn{1}{c}{\textbf{30.9\%}}  \\ 

   \midrule

 \multirow{6}{*}{ThreeOP} & 
 \multirow{2}{*}{{SD V1.4}} &
  \textit{CR} &
\multicolumn{1}{c}{12.4\%} & 
\multicolumn{1}{c}{28.6\%} & 
  \multicolumn{1}{c}{32.2\%} & 
  \multicolumn{1}{c}{29.0\%} & 
  \multicolumn{1}{c}{\textbf{41.0\%}} \\ 

  &  & \textit{CLIPScore} &
\multicolumn{1}{c}{31.2\%} & 
\multicolumn{1}{c}{31.3\%} & 
  \multicolumn{1}{c}{29.7\%} & 
  \multicolumn{1}{c}{31.6\%} & 
  \multicolumn{1}{c}{\textbf{31.7\%}}  \\ 

\cline{2-8}
  \rule{0pt}{2.6ex}
  
& \multirow{2}{*}{{SD V1.5}} &
  \textit{CR} &
\multicolumn{1}{c}{13.4\%} & 
\multicolumn{1}{c}{28.9\%} & 
  \multicolumn{1}{c}{32.2\%} & 
  \multicolumn{1}{c}{32.6\%} & 
  \multicolumn{1}{c}{\textbf{46.4\%}} \\ 

  &  & \textit{CLIPScore} &
\multicolumn{1}{c}{31.2\%} & 
\multicolumn{1}{c}{31.3\%} & 
  \multicolumn{1}{c}{30.2\%} & 
  \multicolumn{1}{c}{\textbf{31.6\%}} & 
  \multicolumn{1}{c}{\textbf{31.6\%}}  \\ 

\cline{2-8}
  \rule{0pt}{2.6ex}

& \multirow{2}{*}{{SD V2.1}} &
  \textit{CR} &
\multicolumn{1}{c}{14.0\%} & 
\multicolumn{1}{c}{30.1\%} & 
  \multicolumn{1}{c}{33.6\%} & 
  \multicolumn{1}{c}{34.3\%} & 
  \multicolumn{1}{c}{\textbf{48.2\%}} \\ 

  &  & \textit{CLIPScore} &
\multicolumn{1}{c}{31.3\%} & 
\multicolumn{1}{c}{31.4\%} & 
  \multicolumn{1}{c}{30.4\%} & 
  \multicolumn{1}{c}{31.7\%} & 
  \multicolumn{1}{c}{\textbf{31.8\%}}  \\

  \bottomrule
\end{tabular}
  \end{threeparttable}
}
\end{table}

For the CLIPScore, the results shows that the improvements are subtle.
The reason is that for prompts containing multiple objects, the presence of some objects from the prompt in the generated image can still result in a high CLIPScore.
Therefore, the similarity difference between correctly generated images and incorrectly generated images with respect to the original prompts is subtle.
Furthermore, as we illustrated in Section \ref{sec:metric}, CLIPScore is a weak indicator with which we can not directly infer whether a generated image is correct or not.
By comparison, CR together with the manual judgement is more suitable and direct to evaluate whether the catastrophic-neglect issue is mitigated or not.
In general, the results demonstrate that {\tool} significantly improves CR while maintaining CLIPScore compared to original dataset, demonstrating it's effectiveness.


\subsection{Ablation Study}
\begin{table}[t] 
\setlength{\abovecaptionskip}{5pt}   
  \setlength{\belowcaptionskip}{0pt}
\Huge
  \caption{The Correct Rate of the original prompts, Explicit Feature Enhancing (EFE), Implicit Feature Enhancing (IFE) and {\tool}.}
  \label{tab:RQ2_ablation}
\resizebox{0.5\textwidth }{!}{ \small
\begin{threeparttable}
\begin{tabular}{cc|cccc}
\toprule
\multirow{1}{*}{\textbf{Dataset}} &
\multirow{1}{*}{\textbf{Model}} &
  \multirow{1}{*}{\textbf{Original}} &
  \multirow{1}{*}{\textbf{EFE}} &
   \multirow{1}{*}{\textbf{IFE}} &
   \multirow{1}{*}{\textbf{{\tool}}} \\     

\midrule
\multirow{3}{*}{TBP} &
\multirow{1}{*}{{SD V1.4}} &

  \multicolumn{1}{c}{61.4\%} & 
  \multicolumn{1}{c}{82.6\%} & 
  \multicolumn{1}{c}{79.3\%} & 
  \multicolumn{1}{c}{\textbf{89.8\%}}  \\

\cline{2-6}
  \rule{0pt}{2.6ex}

& \multirow{1}{*}{{SD V1.5}}&
\multicolumn{1}{c}{55.1\%} & 
  \multicolumn{1}{c}{81.0\%} & 
  \multicolumn{1}{c}{74.2\%} & 
  \multicolumn{1}{c}{\textbf{88.0\%}} \\

\cline{2-6}
  \rule{0pt}{2.6ex}

& \multirow{1}{*}{{SD V2.1}} &
\multicolumn{1}{c}{72.4\%} & 
  \multicolumn{1}{c}{90.1\%} & 
  \multicolumn{1}{c}{83.7\%} & 
  \multicolumn{1}{c}{\textbf{96.0\%}} \\

  \midrule

 \multirow{3}{*}{TwOP} &
 \multirow{1}{*}{{SD V1.4}}  &
\multicolumn{1}{c}{45.6\%} & 
  \multicolumn{1}{c}{73.2\%} & 
  \multicolumn{1}{c}{59.0\%} & 
  \multicolumn{1}{c}{\textbf{77.8\%}} \\

\cline{2-6}
  \rule{0pt}{2.6ex}
  
& \multirow{1}{*}{{SD V1.5}} &
\multicolumn{1}{c}{45.8\%} & 
  \multicolumn{1}{c}{70.4\%} & 
  \multicolumn{1}{c}{61.6\%} & 
  \multicolumn{1}{c}{\textbf{78.0\%}} \\

\cline{2-6}
  \rule{0pt}{2.6ex}

& \multirow{1}{*}{{SD V2.1}} &
\multicolumn{1}{c}{49.6\%} & 
  \multicolumn{1}{c}{75.6\%} & 
  \multicolumn{1}{c}{66.4\%} & 
  \multicolumn{1}{c}{\textbf{80.2\%}} \\

   \midrule

 \multirow{3}{*}{ThreeOP} & 
 \multirow{1}{*}{{SD V1.4}}  &
\multicolumn{1}{c}{12.4\%} & 
  \multicolumn{1}{c}{34.0\%} & 
  \multicolumn{1}{c}{24.6\%} & 
  \multicolumn{1}{c}{\textbf{41.0\%}} \\

\cline{2-6}
  \rule{0pt}{2.6ex}
  
& \multirow{1}{*}{{SD V1.5}} &
\multicolumn{1}{c}{13.4\%} & 
  \multicolumn{1}{c}{40.2\%} & 
  \multicolumn{1}{c}{27.2\%} & 
  \multicolumn{1}{c}{\textbf{46.4\%}} \\

\cline{2-6}
  \rule{0pt}{2.6ex}

& \multirow{1}{*}{{SD V2.1}} &
\multicolumn{1}{c}{14.0\%} & 
  \multicolumn{1}{c}{41.8\%} & 
  \multicolumn{1}{c}{29.5\%} & 
  \multicolumn{1}{c}{\textbf{48.2\%}} \\

  \bottomrule
\end{tabular}
  \end{threeparttable}
}
\end{table}
To investigate the effectiveness of the core component in {\tool}, we conducted ablation experiments to explore the Correct Rate (CR) after removing Explicit Feature Enhancement (EFE) and Implicit Feature Enhancement (IFE) individually.
The results, as shown in Table \ref{tab:RQ2_ablation}, show that both EFE and IFE significantly improve CR.
Specifically, for datasets containing prompts with two objects, EFE and IFE achieve CRs of 78.8\% and 70.7\%, respectively, which are 23.9\% and 15.8\% higher than the CR of the original dataset.
For datasets containing prompts with three objects, EFE and IFE achieve CR improvements of 25.4\% and 13.8\%, respectively, compared to the original dataset.
This demonstrates the effectiveness of each component of {\tool}.
Moreover, combining EFE and IFE achieves a higher CR, indicating that the two components complement each other and that their combination can address a broader scope of catastrophic-neglect issue.

\section{Related Work}
\label{sec:related_work}
\subsection{Text-to-Image Diffusion Models}
In recent years, the diffusion model has emerged as a more advanced and popular framework for text-to-image (T2I) generation compared to traditional non-diffusion methods like Variational Autoencoders (VAEs) \cite{YanYSL16,MansimovPBS15} and Generative Adversarial Networks (GANs) \cite{ZhuP0019,YeYTSJ21}. 
Compared to GANs and VAEs, diffusion models achieve better results due to their stability during training and ability to progressively refine images, leading to higher quality and more detailed outputs \cite{HoJA20,NicholD21} .
 To control the generation of diffusion models, 
   \citet{DhariwalN21} firstly propose a conditional image synthesis method utilizing classifier guidance, achieving great success in text-to-image generation.
Following that, some representative studies \cite{Bao2022,Ramesh2022,RombachBLEO22,SahariaCSLWDGLA22} of text-to-image diffusion models have
been proposed, based on the conditioning mechanism.
Our experiments are based on Stable Diffusion \cite{RombachBLEO22} considering its wide applications.

\subsection{Different Issue Types in T2I DM}
With the rapid development of T2I DMs, researchers have primarily focused on two main aspects: safety issue and fundamental performance issue \citet{zhai2023text,zhai2024membership,zhai2024discovering,liu2024groot,borji2023qualitative}.

As for the issue in fundamental performance,
 \citet{borji2023qualitative} systematically discusses all existing issues in image generation but does not analyze the causes of the catastrophic-neglect issue in T2I models when prompts contain multiple object descriptions. According to the motivation, we discovered that catastrophic-neglect is the most prevalent issue (accounts for 94.0\% in Error Rate) when prompts include multiple object descriptions. 
\citet{liu2023discoveringfailuremodestextguided} mentions the issue of object omission, assuming that specific action descriptions cause some objects to be missing in the image. Our proposal addresses object omission caused by inconsistent features among multiple objects, highlighting a different insight.
\citet{samuel2024generating} addresses the issue of text-to-image models generating incorrect objects for rare concepts. It focuses more on single objects, which is not consistent with the issue our approach aims to solve.
\citet{aithal2024understanding} discusses the hallucination issue, where text-to-image generated images contain samples that have never existed in the training set. This type of hallucination issue is not related to the catastrophic-neglect issue we are addressing.

\subsection{Optimizations For T2I DM}

Some research efforts have focused on optimizing the inference process of T2I DMs. For instance, \citet{LiuLDTT22, FengHFJANBWW23, CheferAVWC23} have worked on improving the guidance mechanism through cross-attention, enabling T2I DMs to better focus on each object and attribute within the prompts, which helps in generating more accurate images.
Additionally, there are works focusing on hand-crafted guidelines for prompt optimization. These studies involve selecting and composing prompts to generate images that achieve a distinct visual style and high quality \cite{LiuC22a, oppenlaender2022taxonomy}. 
Such approaches often rely on manual intervention and expert knowledge.
To automate the construction of optimized prompts, \citet{HaoC0W23} propose an approach that combines supervised learning and reinforcement learning to train a prompt optimization model. The optimized prompts generated by this model are able to produce more aesthetically pleasing images and better adhere to the semantic content of the prompts.
Just as large language models exhibit biases in their understanding of different words \citet{li2024glitch}, T2I DMs face similar issues. This leads to the issue of unbalanced object characteristics when describing multiple objects.
In this study, we focus on repairing catastrophic-neglect in T2I DMs by optimizing at the prompt level.



\section{Conclusion}
\label{sec:conclusion}
This paper proposes an approach (\tool) to repair catastrophic-neglect in Text-to-Image Diffusion Models by attention-guided features enhancement of neglected objects in the generated images.
{\tool} first inputs the prompt into a T2I DM and an attention explainability tool to obtain the generated image and the attention scores for each token.
It then checks whether all objects in the prompt appear in the generated image based on the text-image similarity.
If any objects are neglected, {\tool} iteratively searches for suitable explicit and implicit features to enhance the neglected objects based on the attention differences between the objects.
Experimental results demonstrate that {\tool} effectively addresses the issue of catastrophic-neglect in T2I DMs, achieving a 10.1\%-16.3\% higher Correct Rate based on manual annotation compared to baselines, as tested on Stable-Diffusion V1.4, V1.5, and V2.1 models. 
Additionally, ablation experiments show that both explicit feature enhancing and implicit feature enhancing in {\tool} contribute to resolving the issue of catastrophic-neglect in T2I DMs.

\section*{Limitations}
\label{sec:Limitation}
There are two limitations to the current study.
Firstly, {\tool} primarily repair the issue of catastrophic-neglect for objects and does not consider errors related to the attributes of the objects.
Considering that attribute repairing requires the accurate generation of objects as a foundation and the catastrophic-neglect is a prevalent issue in T2I DMs, this work first attempts to repair object neglect.
We will explore the repair of attribute neglect in future work.

Secondly, {\tool} requires multiple iterations to identify suitable features for generating correct repaired prompts, which increases the time cost of repair.
To mitigate this, we utilize attention differences to guide the search for the optimal features, and results show that, on average, only 2.3 iterations are needed to find the enhanced features that can correctly repair the image.
\bibliographystyle{acl_natbib}
\bibliography{ref}
\appendix
\section{Details of Explicit Feature Prompts}
\label{sec:appendix_prompt}
The details of the explicit feature prompts are illustrated below.
In {\tool}, we replace the placeholders in the following prompts with the neglected objects.

\begin{tcolorbox}
\small
\textbf{Shape Feature Prompt:}

What are the common shapes of the \textit{\textbf{[Neglected Object]}}?

Please output the answer without explanation.
There are two guidelines:
1) The output should add shapes to the neglected object to construct a fluent phrase, separating each phrase with a semicolon;
2) Each shape should originate from a distinct perspective.

\textbf{Example:}

\textbf{Question:} What are the common shapes of bicycle?

\textbf{Output:} two-wheeled bicycle; bicycle with pedals; bicycle with chain and gears
\end{tcolorbox}

\begin{tcolorbox}
\small
\textbf{Color Feature Prompt:}

What are the most common color of the \textit{\textbf{[Neglected Object]}}?

Please output the answer without explanation.
There are two guidelines:
1) The output should add colors to the neglected object to construct a fluent phrase, separating each phrase with a semicolon;
2) Each color should originate from a distinct perspective.

\textbf{Example:}

\textbf{Question:} What are the most common colors of apple?

\textbf{Output:} red apple; green apple
\end{tcolorbox}

\section{Details of The Prompt in LLM-Repair}
\label{sec:LLMRepair}
The details of the prompt in LLM-Repair is illustrated below. In Patcher, we replace the placeholders in the following prompts with the input prompt and the neglected object.

\begin{tcolorbox}
\small
\textbf{LLM-Repair Prompt:}

Input Prompt: \textit{\textbf{[Input Prompt]}}

The input prompt is fed into the Text-to-Image model.
However, the \textit{\textbf{[Neglected Object]}} is not shown on the generated image.
Please repair the input prompt and output eight repaired prompt and and separating each prompt with a semicolon without explanation.
\end{tcolorbox}




\section{Examples of Images Generated by Original Prompts and Repaired Prompts Derived from {\tool}}
\begin{figure}[!t]
\centering
\setlength{\abovecaptionskip}{5pt}   
  \setlength{\belowcaptionskip}{0pt} 
\includegraphics[width=9cm]{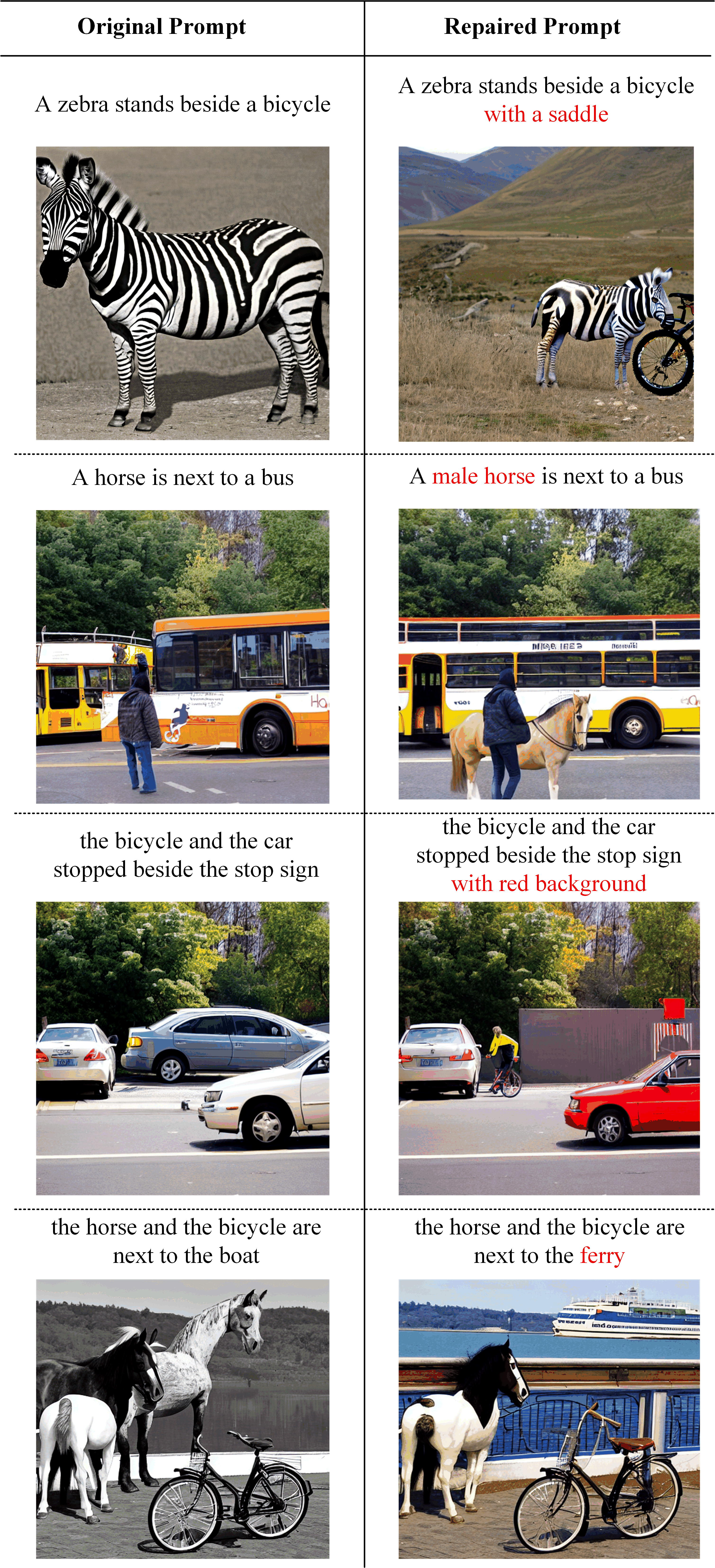}
\caption{
 Images generated by original prompts and repaired prompts.
}
\label{fig:appendix}
\end{figure}

Examples of the images generated from the original prompt and the repaired prompt generated by {\tool} are shown in the figure \ref{fig:appendix}.

\end{document}